\documentclass{article}
\usepackage{spconf,amsmath,graphicx}
\usepackage{amssymb}
\usepackage[utf8]{inputenc}
\usepackage{newunicodechar}
\usepackage{multirow}
\usepackage{enumitem}
\setlist{nosep, leftmargin=14pt}

\usepackage{mwe} 

\title{Interpretable Graph Convolutional Network of multi-modality brain imaging for Alzheimer's disease diagnosis}
\name{Houliang Zhou$^{1}$, Lifang He$^{1}$, Yu Zhang$^{2}$, Li Shen$^{3}$, Brian Chen$^{1}$}

\address{$^{1}$ Department of Computer Science and Engineering, Lehigh University, PA, USA\\
$^{2}$ Department of Bioengineering, Lehigh University, PA, USA\\
$^{3}$ Department of Biostatistics, Epidemiology and Informatics, University of Pennsylvania, PA, USA}
\begin{document}
%
\maketitle
\begin{abstract}
Identification of brain regions related to the specific neurological disorders are of great importance for biomarker and diagnostic studies. In this paper, we propose an interpretable Graph Convolutional Network (GCN) framework for the identification and classification of Alzheimer's disease (AD) using multi-modality brain imaging data.  Specifically, we extended the Gradient Class Activation Mapping (Grad-CAM) technique to quantify the most discriminative features identified by GCN from brain connectivity patterns. We then utilized them to find signature regions of interest (ROIs) by detecting the difference of features between regions in healthy control (HC), mild cognitive impairment (MCI), and AD groups. We conducted the experiments on the ADNI database with imaging data from three modalities, including VBM-MRI, FDG-PET, and AV45-PET, and showed that the ROI features learned by our method were effective for enhancing the performances of both clinical score prediction and disease status identification. It also successfully identified biomarkers associated with AD and MCI. 

\end{abstract}
%

\section{Introduction}
\label{sec:intro}

Neuroimaging pattern classification methods have demonstrated recent advances in predicting Alzheimer’s disease (AD) and mild cognitive impairment (MCI) from magnetic resonance imaging (MRI) and positron emission tomography (PET) scans \cite{wang2011sparse,du2019multi,izquierdo2017robust}. Since the brain is an extremely complex system, large improvements in understanding the brain's organization have been made by representing the brain as a connectivity graph \cite{rykhlevskaia2008combining}. In this graph, nodes are defined as brain regions of interest (ROIs) and edges are defined as the connectivity between those ROIs. This representation is highly compatible with Graph Convolutional Network (GCN), a deep learning method with demonstrated capabilities for analyzing graph structure problems \cite{zhang2018end,kipf2016semi,duvenaud2015convolutional,gilmer2017neural}. \\
\indent In neuroimaging, GCN has been widely used to analyze the brain connectivity graph and discover neurological biomarkers \cite{li2021braingnn,cui2021brainnnexplainer}. In the graph classification problem, the explainability of GCN predictions is crucial for helping to identify and localize biomarkers that contribute to AD or MCI. Several approaches have been proposed to explain the GCN model (e.g., \cite{ying2019gnnexplainer,luo2020parameterized,vu2020pgm}). Most of them did not target disease prediction. Generally, recent methods for interpreting brain networks focused only on single modality data \cite{li2021braingnn,cui2021brainnnexplainer}. Because we assume that the subjects having the same disease may share the similar patterns in brain network, group-level explanations are significant for identifying salient ROIs and discovering biomarkers related to AD and MCI. 
Recent studies indicated that different imaging modalities can provide essential complementary information that can improve accuracy in disease diagnosis \cite{zhang2018multi}. 
Thus, we consider here the gradient-weighted Class Activation Mapping method (Grad-CAM) for multi-modality imaging data, which has been demonstrated to produce explanations on graphs of moderate size with high fidelity, contrastivity, and sparsity \cite{pope2019explainability}. \\
\indent In this paper, we propose an interpretable GCN framework for the identification and classification of Alzheimer's Disease (AD) using multi-modal brain imaging data. We extended the Grad-CAM technique to interpret the salient ROIs for each subject in HC, AD, and MCI categories. The major contributions of this work include the adaptation of Grad-CAM to the GCN model for interpreting salient ROIs, the integration of three modalities of imaging data to construct the brain connectivity graph, and the extension of the GCN model to discover biomarkers in AD and MCI. In our experimental results, our method exhibited high classification performance, and our interpretation method revealed that subjects in the same disease group share the similar patterns. We also found that the putamen and pallidum biomarkers were important for distinguishing HC, AD and MCI. The features learned by our method have a high correlation with the clinical test scores of Mini-Mental State Examination (MMSE) and Alzheimer's Disease Assessment Score 13 (ADAS13), suggesting that our learned features could potentially predict clinical test scores. These results point to applications for our method in the interpretation of ROIs from imaging data in multiple modalities and disease conditions.


\section{Methods}
\label{sec:method}
\vspace{-.1in}
\subsection{Data Acquisition and Preprocessing}
In this work, we used the image data obtained from Alzheimer's Disease Neuroimaging Initiative (ADNI) \cite{mueller2005alzheimer}. The neuroimaging data were from 755 non-Hispanic Caucasian participants, including 182 HC subjects, 476 MCI subjects, and 97 AD subjects,
which consists of three modalities including structural Magnetic Resonance Imaging (VBM-MRI), fluorodeoxyglucose Positron Emission Tomography (FDG-PET), and 18-F florbetapir PET (AV45-PET).

The multi-modality imaging data were aligned to each participant's same visit. The structural MRI scans were preprocessed with voxel-based morphometry (VBM) using the SPM software \cite{ashburner2000voxel}. Generally, all scans were aligned to a T1-weighted template image, segmented into gray matter (GM), white matter (WM) and cerebrospinal fluid (CSF) maps, normalized to the standard Montreal Neurological Institute (MNI) space as $2 \times 2 \times 2$ mm$^3$ voxels, and were smoothed with an $8$ mm FWHM kernel. The FDG-PET and AV45-PET scans were also registered to the same MNI space by SPM. We subsampled the whole brain and obtained 90 ROIs (excluding the cerebellum and vermis) based on the AAL atlas \cite{tzourio2002automated}. ROI-level measures were calculated by averaging all the voxel-level measures within each ROI.

\subsection{Brain Graph Construction}
To construct the brain connectivity graph from the three modalities, we concatenate all of three modalities into the feature vector for each ROI.  We use 90 ROIs for each subject. Each ROI in the image views as the node in the graph, which can be represented as an undirected weighted graph $G=(V,E)$. The vertex set $V=\{v_1,\cdots,v_n\}$ consists of ROIs in the brain and each edge in $E$ is weighted by a connectivity strength, where $n$ is the number of ROIs. In our work, we define $\widetilde{G}=(V,\widetilde{E})$ based on the multi-modal information of the region using $K$-Nearest Neighbor ($K$NN) graph \cite{zhang2018multi}. The edges are weighted by the Gaussian similarity function of Euclidean distances, i.e., $e(v_i,v_j)=\text{exp}(-\frac{ \lVert v_i-v_j\rVert^2 }{2\sigma^2})$. We identify $N_j$ as the set of $K$-Nearest Neighbors of vertex $v_i$, and connect $v_i$ and $v_j$ if $ v_i \in N_j$ or $ v_j \in N_i$. The weighted adjacency matrix $A \in \mathbb{R}^{N \times N}$ represents the similarity between each ROI and its nearest similar neighbor ROIs. The element of adjacency matrix $A$ can be represented as follows:
\vspace{-.05in}
\begin{equation}
{A}_{i,j}=
\begin{cases}
e(v_i,v_j),& \text{ if $ v_i \in N_j$ or $ v_j \in N_i$ } \\
0,& \text{ otherwise. }
\end{cases}
\end{equation}

\subsection{GCN Model}
The classification of graphs can be achieved by embedding node features into a low dimensional space, grouping nodes, and summarizing them \cite{kipf2016semi}. The summarized vector for each graph is fed into a multilayer perceptron (MLP) classifier. In this work, the whole architecture contains three types of layers, including graph convolutional layers, a node pooling layer, and a readout layer. The graph convolutional layer inductively learns a node representation by recursively transforming and aggregating the feature vectors of its neighboring nodes.

We define a brain adjacency matrix $A \in \mathbb{R}^{N \times N}$ and node feature matrix $X \in \mathbb{R}^{N \times d_{in}} $, where $N$ is the number of ROIs and $d_{in}$ is the dimension of multi-modality input feature. The propagation of GCN model or the forward-pass update of node representation is calculated as:

\vspace{-.05in}
\begin{equation}
H^{l+1}=\sigma(\widetilde{D}^{-\frac{1}{2}} \widetilde{A} \widetilde{D}^{-\frac{1}{2}}H^{l}W^{l})
\end{equation}
\vspace{-.05in}
\\
where $H^0=X$, $H^{l} \in \mathbb{R}^{N \times d_{l}}$ is the output of the $l^{th}$ graph convolution layer, $d_l$ is the number of  output channels of layer $l$,  $\widetilde{A} = A+I$ is the adjacency matrix of graph with self -loops, $I\in \mathbb{R}^{N \times N}$ is the identity matrix, $W^l\in \mathbb{R}^{d_l \times d_{l+1}}$ are the learnable parameters, and $\widetilde{D}$ is the diagonal degree matrix with  
$\widetilde{D}_{i,i} = \sum_{j}\widetilde{A}_{i,j}$.
We normalize the $\widetilde{A}$ by multiplying $\widetilde{D}^{-\frac{1}{2}}$ in order to keep a fixed feature scale after graph convolution. $\sigma$ is the activation function.

The node pooling layer groups the nodes together to summarize the features of whole graph. After we get the final output $H^L$ of graph convolution layer, we will summarize the whole feature matrix $H^L$ into a single vector, which is fed into an MLP classifier with softmax activation function in the final layer. We use the negative log-likelihood as the loss function for graph classification.

\subsection{Interpretability of GCN}
We now present how we can use the above formulation for explaining GNN predictions and identifying important ROIs. Our idea draws inspiration from the recent work on GNN explainability using Gradient Class Activation Mapping (Grad-CAM) \cite{pope2019explainability}, which was originally proposed for producing visual explanations for CNN models. We first define the $k^{th}$ graph convolutional feature map at layer $l$ as:

\vspace{-.05in}
\begin{equation}
H_k^{l+1}=\sigma(\widetilde{D}^{-\frac{1}{2}} \widetilde{A} \widetilde{D}^{-\frac{1}{2}}H^{l}W_k^{l})
\end{equation}
\vspace{-.05in}
\\
where $W_k^{l}$ is the $k^{th}$ column of learnable matrix $W^{l}$. According to this notation, we denote $H_{k,n}^{l}$ is the $k^{th}$ feature at the $l^{th}$ layer for each node $n$. We denote $L$ as the final GCN layer. $H_{k,n}^{L}$ represents the feature map of the final convolutional layer.

The $k^{th}$ average feature map after the final convolution layer and the class score for each class $c$ can be calculated as:

\vspace{-.2in}
\begin{equation}
\begin{split}
e_k=\frac{1}{N}\sum_{n=1}^{N}H_{k,n}^L\\
y^c=\sum_k w_k^c e_k
\end{split}
\end{equation}
\vspace{-.1in}
\\
where the learnable $w_k^c$ encodes the $k^{th}$ feature importance for predicting the class $c$, and $y^c$ is the class score. The class specific weights $\beta_k^c$ of Grad-CAM for class $c$ and feature $k$ at final layer $L$ can be calculated as:

\vspace{-.1in}
\begin{equation}
\beta_k^c = \frac{1}{N}\sum_{n=1}^{N} \frac{\partial y^c}{\partial H_{k,n}^L}
\end{equation}
\vspace{-.1in}

The Grad-CAM's heat map can be calculated at the final convolutional layer as:

\vspace{-.1in}
\begin{equation}
L_n^c = ReLU(\sum_k \beta_k^c H_{k,n}^L)
\label{eq:heat-map}
\end{equation}
\vspace{-.1in}
\\
where $L_n^c$ is the heat map of node $n$ for class $c$ at the final GCN layer. We apply the equation (\ref{eq:heat-map}) to get the node importance for each ROI in the brain connectivity graph.

\vspace{-.1in}
\section{Results}
\label{sec:result}

\vspace{-.1in}
\subsection{Classification Performance}
\label{sec:result_classify}
In our experiments, we separate the whole data into three groups including HC vs AD, HC vs MCI, and MCI vs AD to examine our classification results. We use the one vs one strategy to classify three groups and compare the average results when using one, two or three modalities.  This experimental design measures the performance of our method on different numbers of modalities, since it is a multimodal method, and on different groups.

For the brain connectivity graphs, we used $K=10$ to build the $K$NN graph for each subject. After building the graph data, we used the GCN + MLP to calculate the classification result with the following configuration: three graph convolutional layers of size 10, 10, and 5, respectively, followed by three fully-connected layers, a dropout layer, and a softmax classifier. The models were trained for 100 epochs using the Adam optimizer with $\beta_1=0.9$, $\beta_2=0.999$, and a learning rate of 0.005, where the parameters $\beta_1$ and $\beta_2$ control the exponential decay rates of the moving averages of gradients. The dropout rate is 0.5. We performed 5-fold cross validation and repeated the experiment for 50 times. The average classification accuracy, ROC-AUC, sensitivity, specificity, and their standard deviations are reported.

\begin{figure}[htb]
\centering
\centerline{\includegraphics[width=3.3cm]{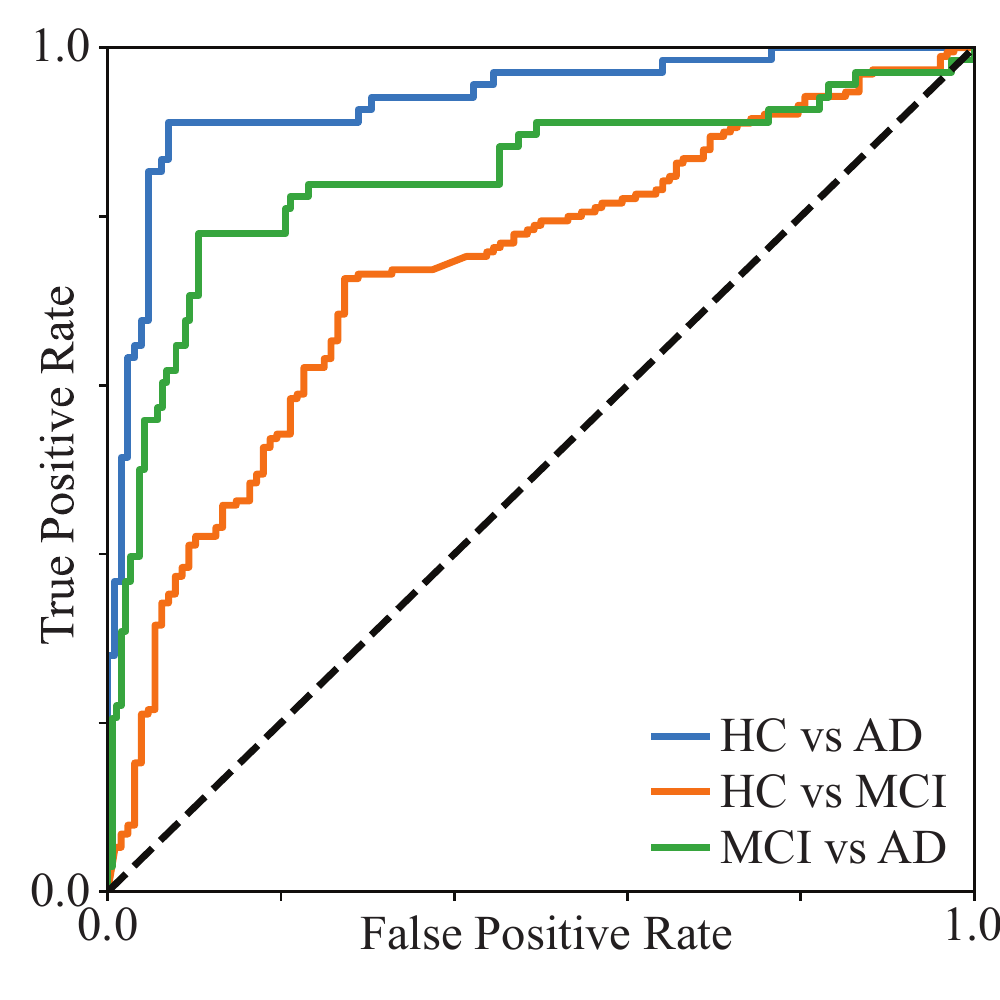}}
\vspace{-.1in}
\caption{ROC Curve of different groups with three modalities.}
\label{fig:classify_auc}
\vspace{-.2in}
\end{figure}

Tables \ref{table:classify_1modal}, \ref{table:classify_2modal} and \ref{table:classify_3modal} show the classification results by using one, two and three modalities, respectively, where $\pm$ represents the standard deviation of evaluation scores in 5 folds. We compared our method with SVM (RBF kernel), Random Forest (RF), MLP and CNN models, which take vectorized adjacency matrices as inputs. The best classification performance was achieved by our method when using all three modalities. In testing, we concatenated the predicted possibilities and the labels of testing data in 5 folds to build the ROC curve by calculating the true positive rate and false positive rate. The ROC curve, using all three modalities, is plotted in Fig. \ref{fig:classify_auc}.  Based on these findings, all three modalities were used to evaluate the interpretability of our model.

\begin{table}[]
  \centering
  \caption{\textbf{One Modality}. Classification results of different groups with VBM-MRI only.}
  \label{table:classify_1modal}
  \scalebox{0.85}{
  \begin{tabular}{llllll}
  \hline
   Method & Accuracy & ROC-AUC & Sensitivity  & Specificity \\ \hline
   SVM&	{.723 $\pm.045$} &	.753 $\pm.061$&	.522 $\pm.073$&	.701 $\pm.078$\\
   RF&.761 $\pm.028$&	.751 $\pm.029$&	.568 $\pm.043$&	.669 $\pm.039$\\
   MLP&.734 $\pm.062$&	.742 $\pm.049$&	.615 $\pm.078$&	.665 $\pm.036$\\
   CNN&.769 $\pm.035$&	.669 $\pm.115$&	.611 $\pm.118$&	.652 $\pm.084$\\
   \hline
   Ours&	{.773 $\pm.038$} &	.755 $\pm.045$&	.706 $\pm.057$&	.759 $\pm.046$\\
   \hline
  \end{tabular}}
\vspace{-.1in}
\centering
  \caption{\textbf{Two Modalities}. Classification results of different groups with VBM-MRI and FDG-PET.}
  \label{table:classify_2modal}
  \scalebox{0.85}{
  \begin{tabular}{llllll}
  \hline
   Method & Accuracy & ROC-AUC & Sensitivity  & Specificity \\ \hline
   SVM&	{.746 $\pm.043$} &	.722 $\pm.054$&	.443 $\pm.118$&	.783 $\pm.048$\\
   RF&.771 $\pm.041$&	.795 $\pm.036$&	.591 $\pm.076$&	.679 $\pm.074$\\
   MLP&.795 $\pm.049$&	.773 $\pm.057$&	.651 $\pm.051$&	.668 $\pm.046$\\
   CNN&.783 $\pm.037$&	.689 $\pm.085$&	.597 $\pm.066$&	.662 $\pm.067$\\
   \hline
   Ours&	{.806 $\pm.032$} &	.791 $\pm.052$&	.718 $\pm.061$&	.814 $\pm.043$\\
   \hline
  \end{tabular}}
  
\vspace{-.1in}
\centering
  \caption{\textbf{Three Modalities}. Classification results of different groups with three modalities.}
  \label{table:classify_3modal}
  \scalebox{0.85}{
  \begin{tabular}{llllll}
  \hline
   Method & Accuracy & ROC-AUC & Sensitivity  & Specificity \\ \hline
   SVM&	{.753 $\pm.053$} &	.788 $\pm.042$&	.631 $\pm.087$&	.767 $\pm.079$\\
   RF&.786 $\pm.033$&	.811 $\pm.034$&	.610 $\pm.054$&	.682 $\pm.027$\\
   MLP&.803 $\pm.047$&	.765 $\pm.041$&	.669 $\pm.043$&	.678 $\pm.058$\\
   CNN&.806 $\pm.036$&	.679 $\pm.063$&	.624 $\pm.102$&	.685 $\pm.049$\\
   \hline
   Ours&	{.818 $\pm.031$} &	.815 $\pm.035$&	.747 $\pm.059$&	.821 $\pm.047$\\
   \hline
  \end{tabular}}
  
\vspace{-.1in}  
\end{table}

\subsection{Interpretation of Salient ROIs}
\label{sec:result_Interpretation}
After training models for the ADNI datasets, to summarize the salient ROIs, we apply the Grad-CAM method on all subjects for each class and obtain a set of scalar scores for each ROI, which is the heatmap. For the heatmap of each subject, we rank the scores in descending order and only keep the top 20 salient ROIs. We used the BrainNet Viewer \cite{xia2013brainnet} to plot those salient ROIs on the brain surface.

\begin{figure}[t]
\begin{minipage}[b]{.48\linewidth}
  \centering
  \centerline{\includegraphics[width=3.5cm]{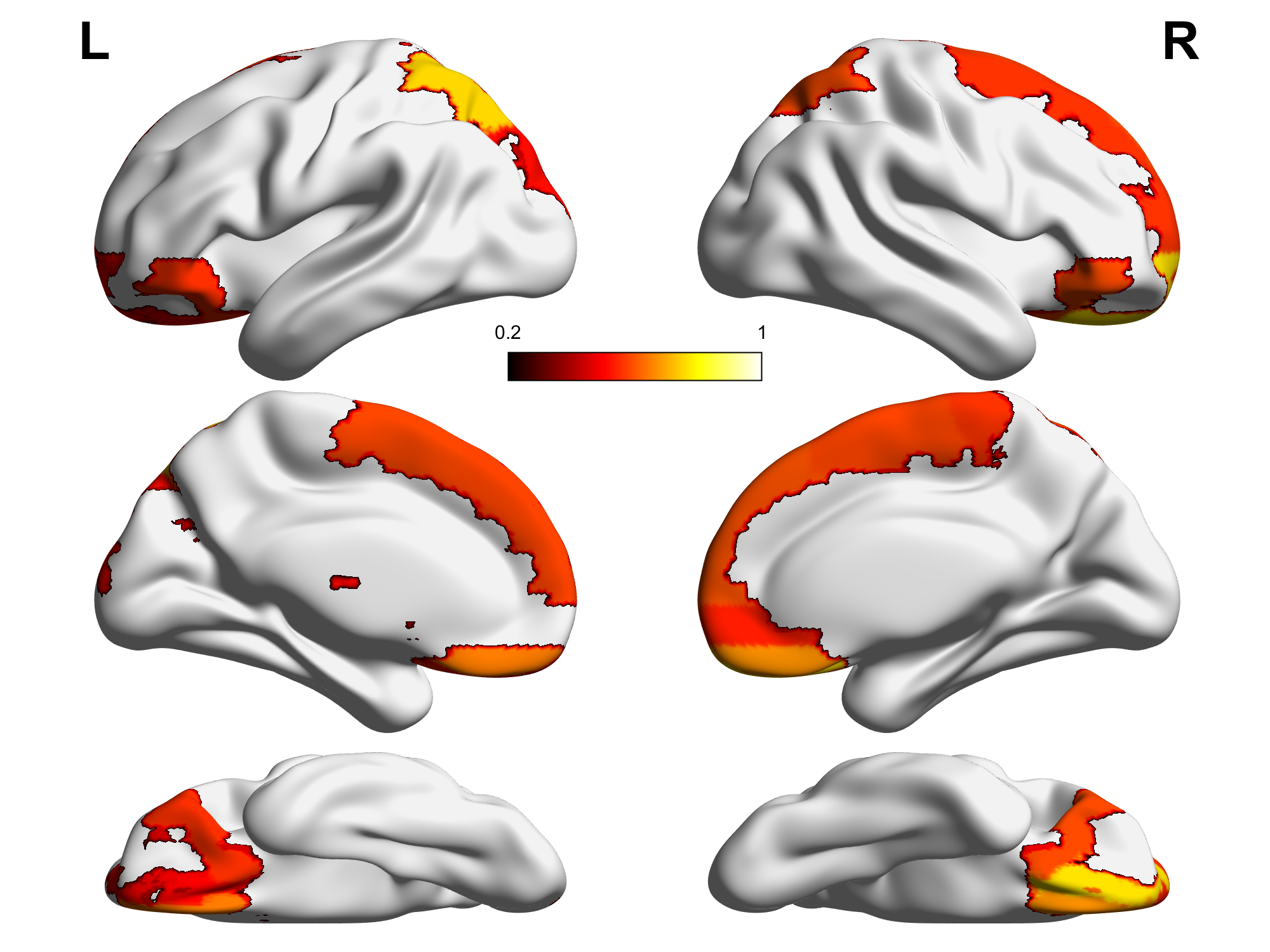}}
\end{minipage}
\hfill
\begin{minipage}[b]{0.48\linewidth}
  \centering
  \centerline{\includegraphics[width=3.5cm]{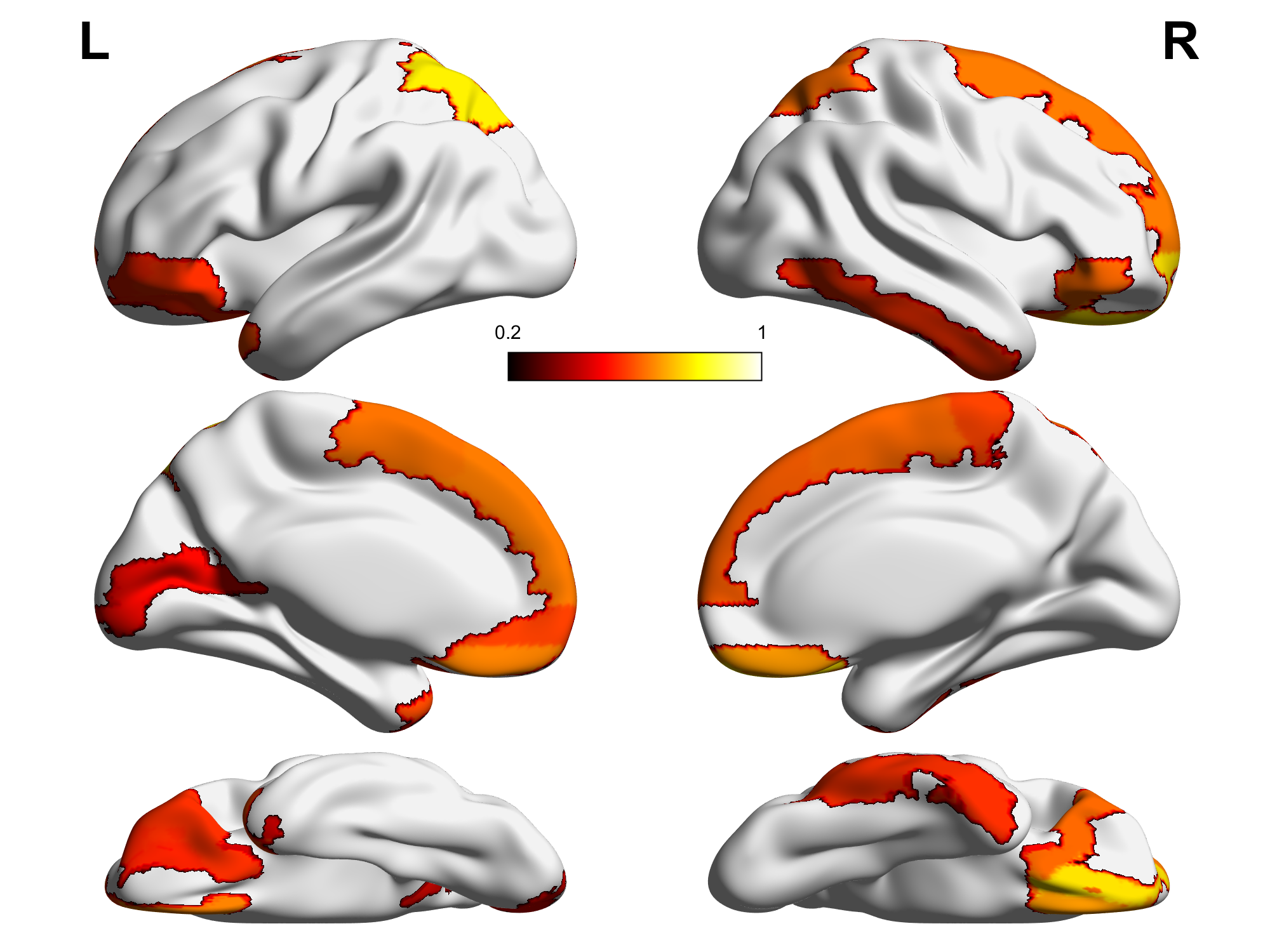}}
\end{minipage}
\centerline{(a) HC Subject}\medskip

\begin{minipage}[b]{.48\linewidth}
  \centering
  \centerline{\includegraphics[width=3.5cm]{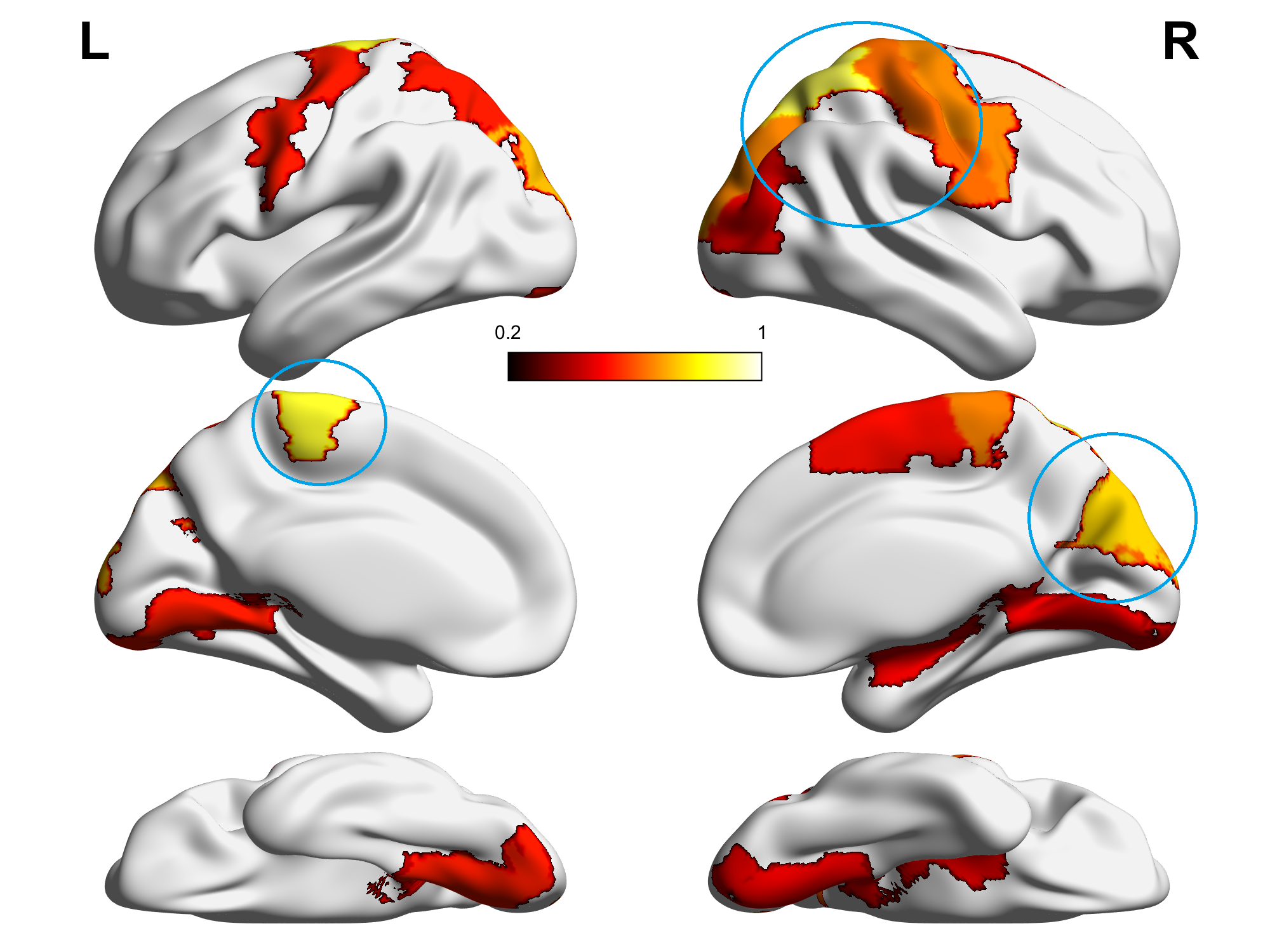}}
\end{minipage}
\hfill
\begin{minipage}[b]{0.48\linewidth}
  \centering
  \centerline{\includegraphics[width=3.5cm]{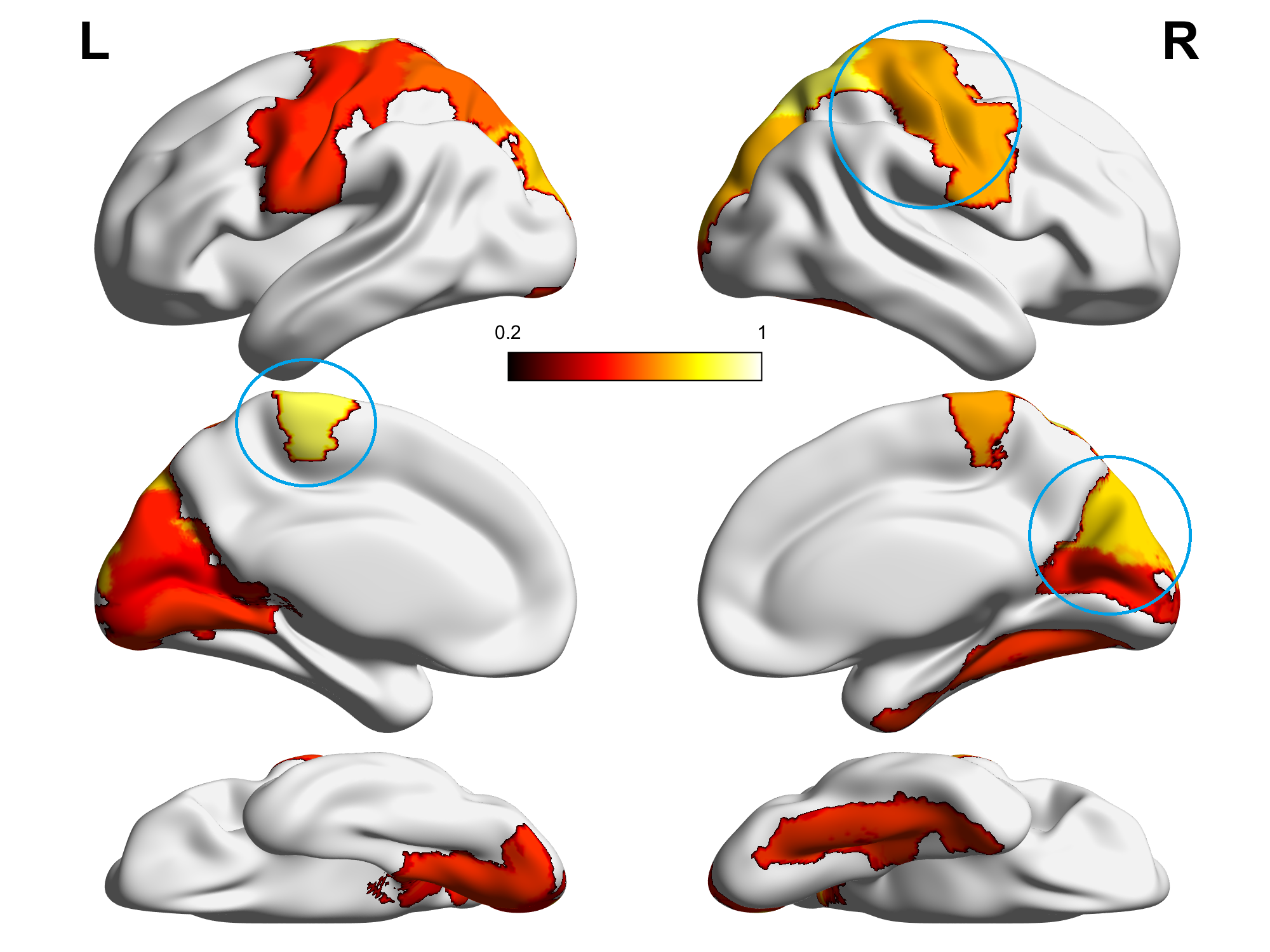}}
\end{minipage}
\centerline{(b) MCI Subject}\medskip

\begin{minipage}[b]{.48\linewidth}
  \centering
  \centerline{\includegraphics[width=3.5cm]{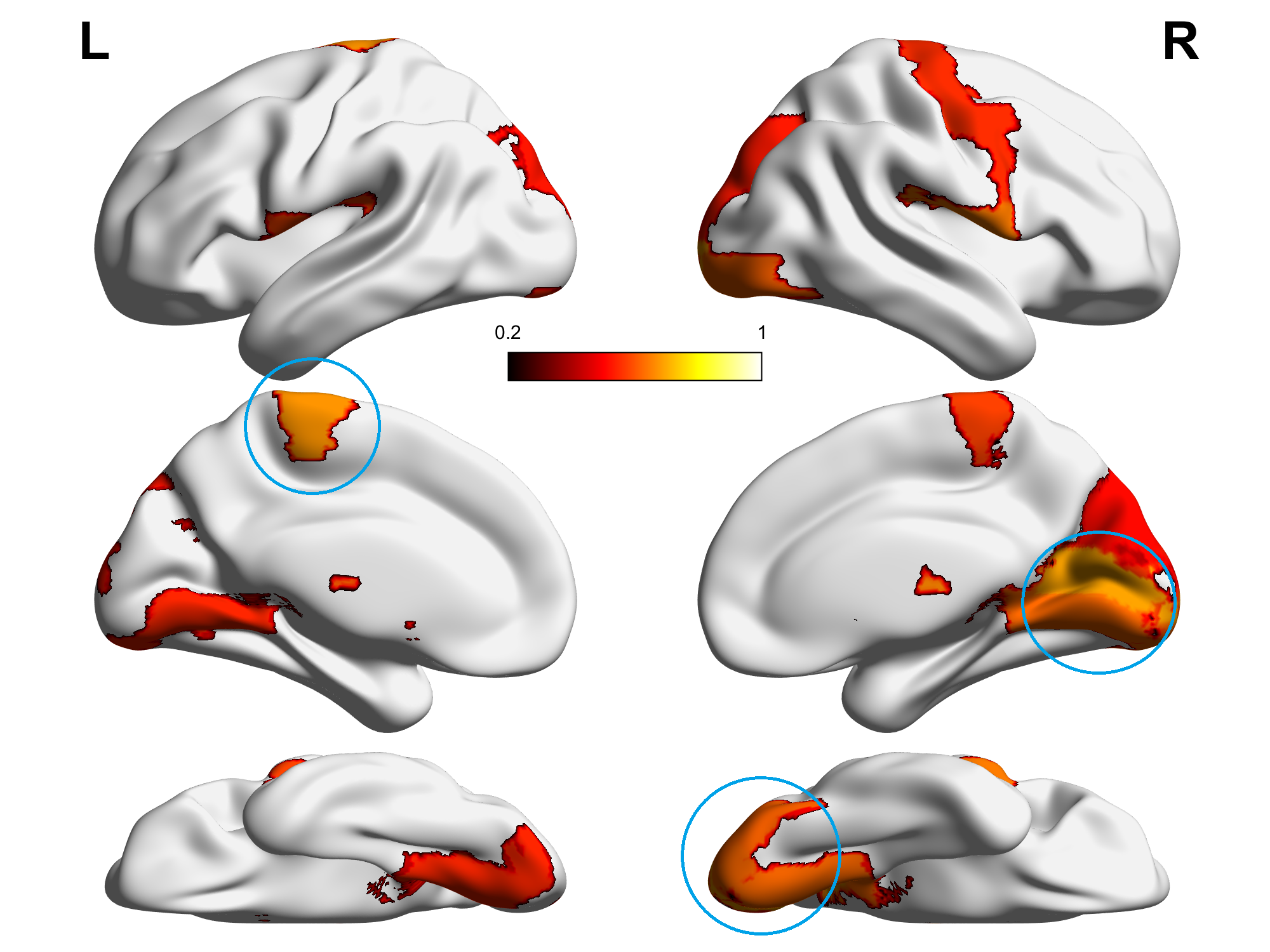}}
\end{minipage}
\hfill
\begin{minipage}[b]{0.48\linewidth}
  \centering
  \centerline{\includegraphics[width=3.5cm]{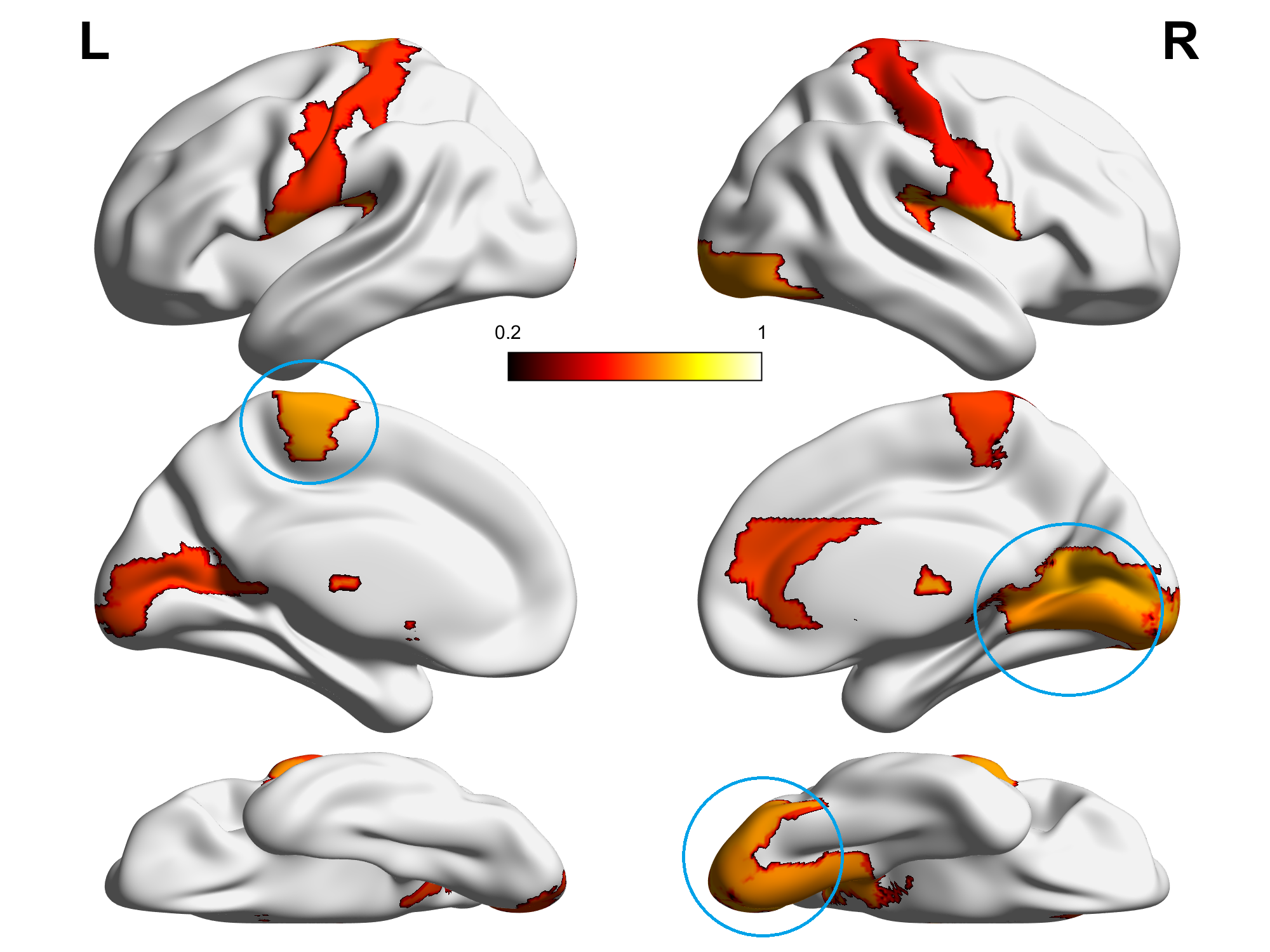}}
\end{minipage}
\vspace{-.1in}
\centerline{(c) AD Subject}\medskip
\vspace{-.1in}
\caption{Interpreting top 20 selected salient ROIs of two different individuals in HC, MCI, and AD group respectively. The color bar ranges from 0.2 to 1.0. The bright-yellow color indicates a high score, while dark-red color indicates a low score. The common detected salient ROIs across different subjects inside MCI and AD are circled in blue.}
\label{fig:interp_rois}
\end{figure}

Fig. \ref{fig:interp_rois}(a-c) illustrates the lateral, medial, and ventral view of brain surface for two subjects selected from the HC, AD, and MCI groups.  We named the selected ROIs as the biomarkers for identifying each group: node importance values were encoded as the color of regions on brain surface. In Fig. \ref{fig:interp_rois}(a), putamen, pallidum, Superior parietal gyrus, and gyrus rectus were selected for HC; In Fig. \ref{fig:interp_rois}(b), putamen, pallidum, Superior occipital and parietal gyrus were selected for MCI; In Fig. \ref{fig:interp_rois}(c), putamen, pallidum, thalamus, Caudate nucleus, Superior occipital and Lingual gyrus were selected for AD. Putamen and pallidum were important for each group. 

\begin{figure}[!t]
\centering
\centerline{\includegraphics[width=3in]{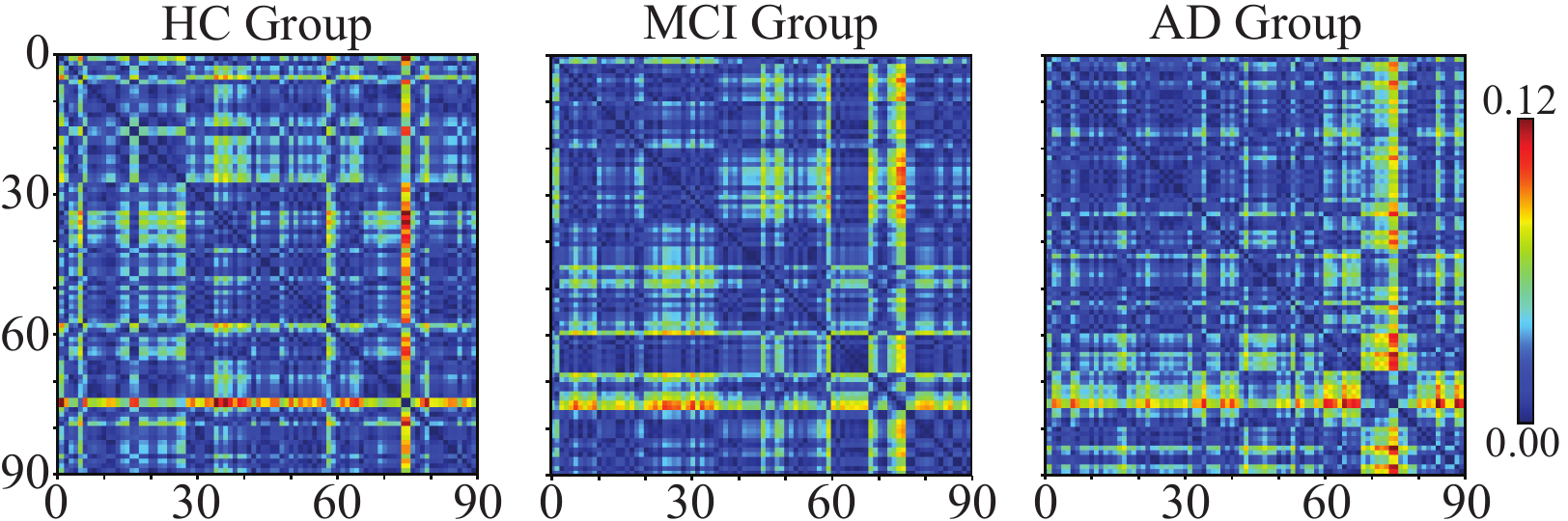}}
\vspace{-.1in}
\caption{The Euclidean distance between ROIs by averaging node importance of subjects inside each group.}
\label{fig:distance_rois}
\vspace{-.1in}
\end{figure}

Given that each ROI represents a node in the graph, after we computed the node importance of each subject, we average those node importance in HC, AD, and MCI group separately. Then we calculate the Euclidean distance between ROIs based on the average node importance for three groups. In Fig. \ref{fig:distance_rois}, we plot the distance matrices between ROIs. The average node importance of HC, AD, and MCI groups were substantially different from each other, suggesting that the salient ROIs are different in each group.

\vspace{-.1in}
\subsection{Prediction Score}
\label{sec:result_score}
In this subsection, we collected the cognitive test scores including the MMSE and ADAS13 from ADNI dataset. We used the learned features of ROIs in the last GCN layer to predict the cognitive tests scores of HC, AD, and MCI subjects. We normalized the learned features as the mean of 0 and the standard deviation of 1. The linear regression model is adopted to fit the relationship between the learned features and the cognitive test scores.

\begin{figure}[!t]
\centerline{\includegraphics[width=2.5in]{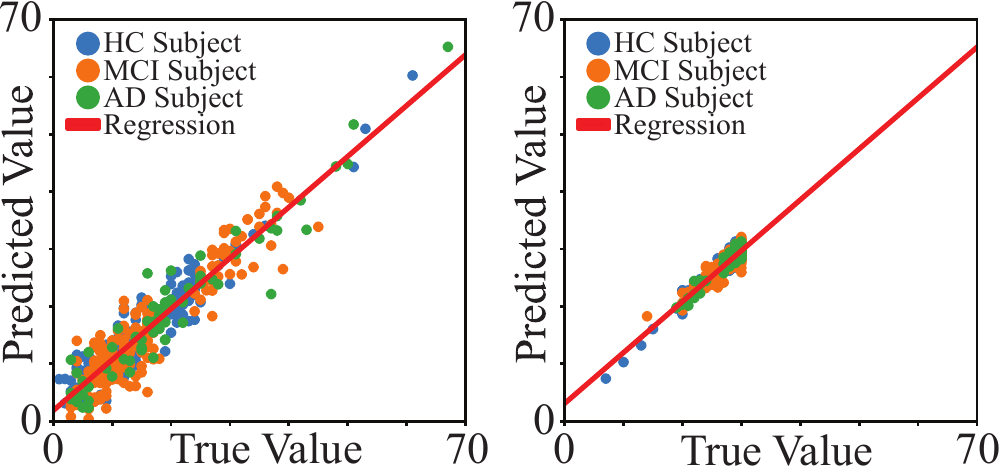}}
\vspace{-.1in}
\caption{Regression between the predicted and true values for ADAS13 and MMSE test scores. The HC, AD, and MCI subjects are plotted as the blue, orange, and green color respectively. The red line is the fitted regression line. }
\label{fig:resgression}
\end{figure}

In Fig. \ref{fig:resgression}, we show the regression measures between the predicted and true values for ADAS13 and MMSE test scores. It provides a visual perception of how accurate the prediction is for the given test. Based on the fit of the regression line, we can infer that there is substantial correlation between the prediction and ground truth.

In Table \ref{table:regression}, we show the numeric performance of the regression results for all of HC, AD, and MCI subjects. It contains the Pearson correlation coefficient, mean absolute error, root mean squared error (RMSE), and R-squared measure for the ADAS13 and MMSE test scores. From the table, the correlation measures of these two cognitive test are higher than 0.9 and the R-squared measures are also very high.

\begin{table}[!t]
\vspace{-.1in}
\centering
  \caption{The reported value of evaluation metrics between the predicted and true scores of ADAS13 and MMSE.}
  \label{table:regression}
  \scalebox{1}{
  \begin{tabular}{lll}
  \hline
   Evaluation Metrics & ADAS13 & MMSE \\ \hline
   Pearson Correlation Coefficient &	0.941&	0.942\\
   Mean Absolute Error	&2.682&	0.848\\
  Root Mean Squared Error	&3.527&	1.084\\
  R Squared &0.886&	0.887\\
   \hline
  \end{tabular}}
\vspace{-.1in}
\end{table}

\vspace{-.1in}
\section{Conclusions}
\label{sec:Conclusions}
\vspace{-.1in}

In this paper, we proposed an interpretable Graph Convolutional Network (GCN) framework for the identification and classification of Alzheimer's Disease (AD) using multi-modality brain imaging data. We applied the interpretable Gradient Class Activation Mapping (Grad-CAM) technique to interpret the salient ROIs and found that the putamen and pallidum biomakers were very important to identify the HC, AD and MCI. We extended the current explanation method of GCN model to discover the neurological biomakers for multi-modal brain imaging analysis. Besides the promising classification performance, our interpretation of salient ROIs demonstrated the individual- and group-level patterns in HC, AD, and MCI groups respectively. This suggests that our method can be applied to interpret the salient ROIs for much more modalities imaging data.

\vfill
\pagebreak

\section{Compliance with Ethical Standards}
\label{sec:ethics}\vspace{-.1in}
This research study was conducted retrospectively using human subject data made available in open access by ADNI \cite{mueller2005alzheimer}. Ethical approval was not required as confirmed by the license attached with the open access data.

\bibliographystyle{IEEEbib}
\bibliography{refs}

\end{document}